\documentclass[11pt,a4paper]{article}
\usepackage[hyperref]{emnlp2018}
\usepackage{times}
\usepackage{latexsym}
\usepackage{amsmath}
\usepackage{CJK}
\usepackage{subfigure}
\usepackage{natbib}
\usepackage{pgfplots}
\usepackage{pgfplotstable}
\pgfplotsset{width=7.5cm,compat=1.5}
\usepackage{color,xcolor,framed}
\usepackage{url}
\usepackage[]{amsfonts}

\aclfinalcopy % Uncomment this line for the final submission
\def\aclpaperid{1235} %  Enter the acl Paper ID here

\newcommand*{\affaddr}[1]{#1} % No op here. Customize it for different styles.
\newcommand*{\affmark}[1][*]{\textsuperscript{#1}}
\newcommand*{\email}[1]{\texttt{#1}}

\title{Semantic-Unit-Based Dilated Convolution for Multi-Label Text Classification}

\author{Junyang Lin\affmark[1,2], Qi Su\affmark[1], Pengcheng  Yang\affmark[2], Shuming Ma\affmark[2], Xu Sun\affmark[2]\\
\affaddr{School of Foreign Languages, Peking University\affmark[1]}\\
\affaddr{MOE Key Lab of Computational Linguistics, School of EECS, Peking University\affmark[2]}\\
\email{\{linjunyang, sukia, yang\_pc, shumingma, xusun\}@pku.edu.cn}\\
}

\date{}

\begin{document}
\maketitle
\begin{CJK}{UTF8}{gbsn}
\begin{abstract}
 We propose a novel model for multi-label text classification, which is based on sequence-to-sequence learning. The model generates higher-level semantic unit representations with multi-level dilated convolution as well as a corresponding hybrid attention mechanism that extracts both the information at the word-level and the level of the semantic unit. Our designed dilated convolution effectively reduces dimension and supports an exponential expansion of receptive fields without loss of local information, and the attention-over-attention mechanism is able to capture more summary relevant information from the source context. Results of our experiments show that the proposed model has significant advantages over the baseline models on the dataset RCV1-V2 and Ren-CECps, and our analysis demonstrates that our model is competitive to the deterministic hierarchical models and it is more robust to classifying low-frequency labels\footnote{The code is available at \url{https://github.com/lancopku/SU4MLC}}.
\end{abstract}

\section{Introduction}

Multi-label text classification refers to assigning multiple labels for a given text, which can be applied to a number of important real-world applications. One typical example is that news on the website often requires labels with the purpose of the improved quality of search and recommendation so that the users can find the preferred information with high efficiency with less disturbance of the irrelevant information. As a significant task of natural language processing, a number of methods have been applied and have gradually achieved satisfactory performance. For instance, a series of methods based on machine learning have been extensively utilized in the industries, such as Binary Relevance \citep{br}. BR treats the task as multiple single-label classifications and can achieve satisfactory performance. With the development of Deep Learning, neural methods are applied to this task and achieved improvements \citep{bp_mll, nn, haram}.

However, these methods cannot model the internal correlations among labels. To capture such correlations, the following work, including ML-DT \citep{ml_dt}, Rank-SVM \citep{rank_svm}, LP \citep{lp}, ML-KNN \citep{ml_knn}, CC \citep{cc}, attempt to capture the relationship, which though demonstrated improvements yet simply captured low-order correlations. A milestone in this field is the application of sequence-to-sequence learning to multi-label text classification \citep{nam2017}. Sequence-to-sequence learning is about the transformation from one type of sequence to another type of sequence, whose most common architecture is the attention-based sequence-to-sequence (Seq2Seq) model. The attention-based Seq2Seq \citep{seq2seq} model is initially designed for neural machine translation (NMT) \citep{attention,stanford_attention}. Seq2Seq is able to encode a given source text and decode the representation for a new sequence to approximate the target text, and with the attention mechanism, the decoder is competent in extracting vital source-side information to improve the quality of decoding. Multi-label text classification can be regarded as the prediction of the target label sequence given a source text, which can be modeled by the Seq2Seq. Moreover, it is able to model the high-order correlations among the source text as well as those among the label sequence with deep recurrent neural networks (RNN).

Nevertheless, we study the attention-based Seq2Seq model for multi-label text classification \citep{nam2017} and find that the attention mechanism does not play a significant role in this task as it does in other NLP tasks, such as NMT and abstractive summarization. In Section~\ref{problem}, we demonstrate the results of our ablation study, which show that the attention mechanism cannot improve the performance of the Seq2Seq model. We hypothesize that compared with neural machine translation, the requirements for neural multi-label text classification are different. The conventional attention mechanism extracts the word-level information from the source context, which makes little contribution to a classification task. For text classification, human does not assign texts labels simply based on the word-level information but usually based on their understanding of the salient meanings in the source text. 

For example, regarding the text ``The young boys are playing basketball with great excitement and apparently they enjoy the fun of competition'', it can be found that there are two salient ideas, which are ``game of the young'' and ``happiness of basketball game'', which we call ``semantic units'' of the text. The semantic units, instead of word-level information, mainly determine that the text can be classified into the target categories ``youth'' and ``sports''.

Semantic units construct the semantic meaning of the whole text. To assign proper labels for text, the model should capture the core semantic units of the source text, the higher-level information compared with word-level information, and then assign the text labels based on its understanding of the semantic units. However, it is difficult to extract information from semantic units as the conventional attention mechanism focuses on extracting word-level information, which contains redundancy and irrelevant details.

In order to capture semantic units in the source text, we analyze the texts and find that the semantic units are often wrapped in phrases or sentences, connecting with other units with the help of contexts. Inspired by the idea of global encoding for summarization \citep{global_encoding}, we utilize the power of convolutional neural networks (CNN) to capture local interaction among words and generate representations of information of higher levels than the word, such as phrase or sentence. Moreover, to tackle the problem of long-term dependency, we design a multi-level dilated convolution for text to capture local correlation and long-term dependency without loss of coverage as we do not apply any form of pooling or strided convolution. Based on the annotations generated by our designed module and those by the original recurrent neural networks, we implement our hybrid attention mechanism with the purpose of capturing information at different levels, and furthermore, it can extract word-level information from the source context based on its attention on the semantic units.

In brief, our contributions are illustrated below: 
\begin{itemize}
\setlength{\itemsep}{0pt}
\setlength{\parsep}{0pt}
\setlength{\parskip}{0pt}
\setlength{\partopsep}{0pt}
\setlength{\topsep}{0pt}
\item We analyze that the conventional attention mechanism is not useful for multi-label text classification, and we propose a novel model with multi-level dilated convolution to capture semantic units in the source text.
\item Experimental results demonstrate that our model outperforms the baseline models and achieves the state-of-the-art performance on the dataset RCV1-v2 and Ren-CECps, and our model is competitive with the hierarchical models with the deterministic setting of sentence or phrase.
\item Our analysis shows that compared with the conventional Seq2Seq model, our model with effective information extracted from the source context can better predict the labels of low frequency, and it is less influenced by the prior distribution of the label sequence.
\end{itemize}

\section{Attention-based Seq2Seq for Multi-label Text Classification}

As illustrated below, multi-label text classification has the potential to be regarded as a task of sequence prediction, as long as there are certain correlation patterns in the label data. Owing to the correlations among labels, it is possible to improve the performance of the model in this task by assigning certain label permutations for the label sequence and maximizing subset accuracy, which means that the label permutation and the corresponding attention-based Seq2Seq method are competent in learning the label classification and the label correlations. By maximizing the subset accuracy, the model can improve the performance of classification with the assistance of the information about the label correlations. Regarding label permutation, a straightforward method is to reorder the label data in accordance with the descending order by frequency, which shows satisfactory effects \citep{cnn-rnn}. 

Multi-label text classification can be regarded as a Seq2Seq learning task, which is formulated as below. Given a source text $x = \{x_1, ..., x_i, ..., x_n\}$ and a target label sequence $y = \{y_1, ..., y_i, ..., y_m\}$, the Seq2Seq model learns to approximate the probability $P(y|x) = \prod_{t=1}^{m}P(y_t|y_{<t}, x)$, where $P(y_t|y_{<t}, x)$ is computed by the Seq2Seq model, which is commonly based on recurrent neural network (RNN).

The encoder, which is bidirectional Long Short-Term Memory (LSTM) \citep{lstm}, encodes the source text $x$ from both directions and generates the source annotations $h$, where the annotations from both directions at each time step are concatenated ($h_i\!=\![\overrightarrow{h_i}; \overleftarrow{h_i}]$). To be specific, the computations of $\overrightarrow{h_i}$ and $\overleftarrow{h_i}$ are illustrated below:
\begin{align}
\overrightarrow{{h_{i}}} = {LSTM}({x_{i}}, \overrightarrow{{h_{i-1}}},{C_{i-1}}) \label{eq7}\\
\overleftarrow{{h_{i}}} = {LSTM}({x_{i}}, \overleftarrow{{h_{i-1}}}, {C_{i-1}}) \label{eq8}
\end{align}

We implement a unidirectional LSTM decoder to generate labels sequentially. At each time step $t$, the decoder generates a label $y_{t}$ by sampling from a distribution of the target label set $P_{vocab}$ until sampling the token representing the end of sentence, where:
\begin{align}
P_{vocab} &= g(y_{t-1}, c_{t},s_{t-1})
\end{align}
where $g(\cdot)$ refers to non-linear functions including the LSTM decoder, the attention mechanism as well as the softmax function for prediction. The attention mechanism generates $c_t$ as shown in the following:
\begin{align}
c_{t} &= \sum^{n}_{i=1} \alpha_{t,i}h_{i}\\
\alpha_{t,i} &= \frac{exp(e_{t,i})}{\sum_{j=1}^{n}exp(e_{t,j})}\\
e_{t,i} &= s_{t-1}^{\top}W_{a}h_{i}
\end{align}

\section{Problem}
\label{problem}
\begin{table}[tb]
		\centering
%         \footnotesize
    	\setlength{\tabcolsep}{5.7pt}
		\begin{tabular}{ l | c c c c }
		\hline
		\multicolumn{1}{ l |}{\textbf{Models}} &
		\multicolumn{1}{c}{\textbf{HL(-)}} & 
		\multicolumn{1}{c}{\textbf{P(+)}} &  
        \multicolumn{1}{c}{\textbf{R(+)}} &
		\multicolumn{1}{c}{\textbf{F1(+)}}   \\  \hline
        w/o attention & 0.0082 & 0.883 & 0.849 & 0.866\\
        +attention & 0.0081 & 0.889 & 0.848 & 0.868\\ 
        \hline
		\end{tabular}
		\caption{Performances of the Seq2Seq models with and without attention on the RCV1-v2 test set, where HL, P, R, and F1 refer to hamming loss, micro-precision, micro-recall and micro-${\rm F_1}$. The symbol “+” indicates that the higher the better, while the symbol “-” indicates that the lower the better.}
		\label{s2s}
\end{table}

As we analyze the effects of the attention mechanism in multi-label text classification, we find that it contributes little to the improvement of the model's performance. To verify the effects of the attention mechanism, we conduct an ablation test to compare the performance of the Seq2Seq model without the attention mechanism and the attention-based SeqSeq model on the multi-label text classification dataset RCV1-v2, which is introduced in detail in Section~\ref{experiment}.

As is shown in Table~\ref{s2s}, the Seq2Seq models with and without the attention mechanism demonstrate similar performances on the RCV1-v2 according to their scores of micro-${\rm F_1}$, a significant evaluation metric for multi-label text classification. This can be a proof that the conventional attention mechanism does not play a significant role in the improvement of the Seq2Seq model's performance. We hypothesize that the conventional attention mechanism does not meet the requirements of multi-label text classification. A common sense for such a classification task is that the classification should be based on the salient ideas of the source text. The semantic units, instead of word-level information, mainly determine that the text can be classified into the target categories ``youth'' and ``sports''. For each of a variety of texts, there are always semantic units that construct the semantic meaning of the whole text. Regarding an automatic system for multi-label text classification, the system should be able to extract the semantic units in the source text for better performance in classification. Therefore, we propose our model to tackle this problem.

\section{Proposed Method}

In the following, we introduce our proposed modules to improve the conventional Seq2Seq model for multi-label text classification. In general, it contains two components: multi-level dilated convolution (MDC) as well as hybrid attention mechanism. 

\subsection{Multi-level Dilated Convolution}
\begin{figure}[tb]
\centering
\includegraphics[width=1.0\linewidth]{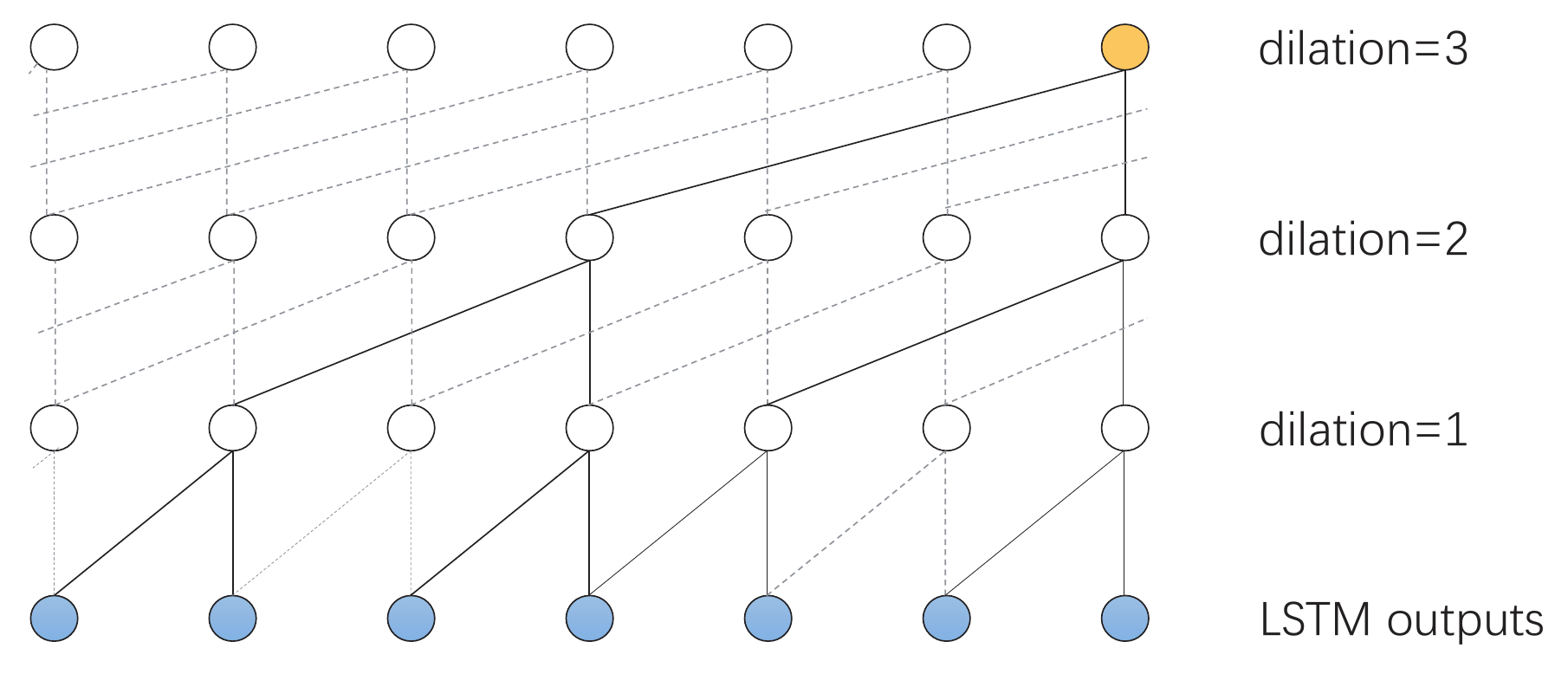}
\caption{\textbf{Structure of Multi-level Dilated Convolution (MDC).} A example of MDC with kernel size $k=2$ and dilation rates $[1,2,3]$. To avoid gridding effects, the dilation rates do not share a common factor other than $1$.}
\end{figure}

On top of the representations generated by the original encoder, which is an LSTM in our model, we apply the multi-layer convolutional neural networks to generate representations of semantic units by capturing local correlations and long-term dependencies among words. To be specific, our CNN is a three-layer one-dimensional CNN. Following the previous work \citep{conv4sent} on CNN for NLP, we use one-dimensional convolution with the number of channels equal to the number of units of the hidden layer, so that the information at each dimension of a representation vector will not be disconnected as 2-dimension convolution does. Besides, as we are to capture semantic units in the source text instead of higher-level word representations, there is no need to use padding for the convolution.

A special design for the CNN is the implementation of dilated convolution. Dilation has become popular in semantic segmentation in computer vision in recent years \citep{dilated_convolution, HDC}, and it has been introduced to the fields of NLP \citep{bytenet} and speech processing \citep{wavenet}. Dilated convolution refers to convolution inserted with ``holes'' so that it is able to remove the negative effects such as information loss caused by common down-sampling methods, such as max-pooling and strided convolution. Besides, it is able to expand the receptive fields at the exponential level without increasing the number of parameters. Thus, it becomes possible for dilated convolution to capture longer-term dependency. Furthermore, with the purpose of avoiding gridding effects caused by dilation (e.g., the dilated segments of the convolutional kernel can cause missing of vital local correlation and break the continuity between word representations), we implement a multi-level dilated convolution with different dilation rates at different levels, where the dilation rates are hyperparameters in our model. 

Instead of using the same dilation rate or dilation rates with the common factor, which can cause gridding effects, we apply multi-level dilated convolution with different dilation rates, such as [1,2,3]. Following \citet{HDC}, for $N$ layers of 1-dimension convolution with kernel size $K$ with dilation rates $[r_1, ..., r_N]$, the maximum distance between two nonzero values is $max(M_{i+1} - 2r_i, M_{i+1}-2(M_{i+1}-r_{i}), r_{i})$ with $M_N = r_N$, and the goal is $M_2 \leq K$. In our experiments, we set the dilation rates to [1, 2, 3] and $K$ to 3, and we have $M_2 = 2$. The implementations can avoid the gridding effects and allows the top layer to access information between longer distance without loss of coverage. Moreover, as there may be irrelevant information to the semantic units at a long distance, we carefully design the dilation rates to [1, 2, 3] based on the performance in validation, instead of others such as [2, 5, 9], so that the top layer will not process the information among overlong distance and reduce the influence of unrelated information. Therefore, our model can generate semantic unit representations from the information at phrase level with small dilation rates and those at sentence level with large dilation rates.

\subsection{Hybrid Attention}

\begin{figure}[t]
\centering
\includegraphics[width=1.0\linewidth]{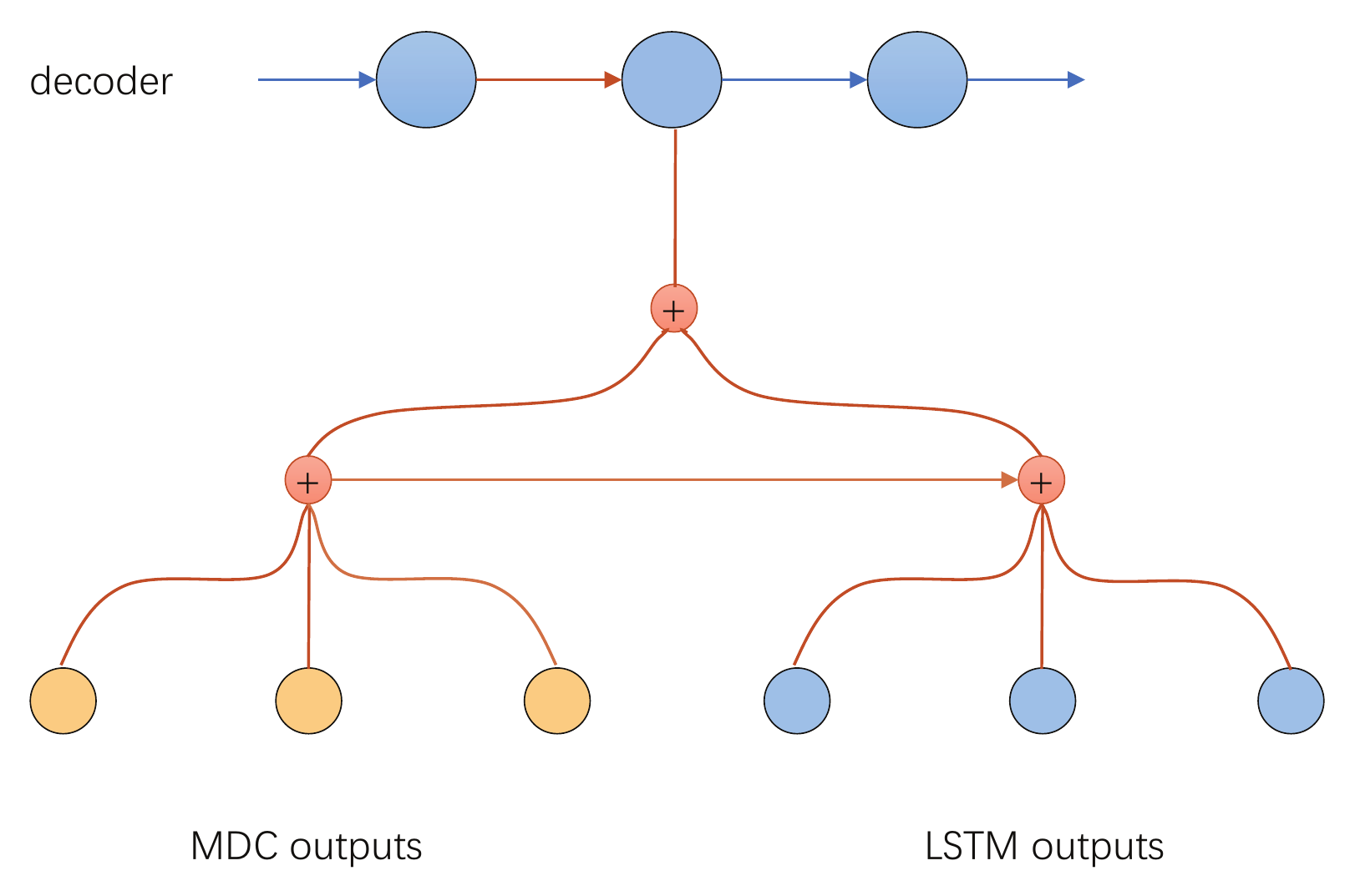}
\caption{\textbf{Structure of Hybrid Attention.} The blue circles at the bottom right represent the source annotations generated by the LSTM encoder, the yellow circles at the bottom left represent the semantic unit representations generated by MDC, and the blue circles at the top represent the LSTM decoder outputs. At each decoding time step, the output of the LSTM attends to the semantic unit representations first, and then the new representation incorporated with high-level information attends to the source annotations.}
\end{figure}

As we have annotations from the RNN encoder and semantic unit representations from the MDC, we design two types of attention mechanism to evaluate the effects of information of different levels. One is the common attention mechanism, which attends to the semantic unit representations instead of the source word annotations as the conventional does, the other is our designed hybrid attention mechanism to incorporate information of the two levels.

The idea of hybrid attention is motivated by memory networks \citep{DBLP:conf/nips/SukhbaatarSWF15} and multi-step attention \citep{fairseq}. It can be regarded as the attention mechanism with multiple ``hops'', with the first hop attending to the higher-level semantic unit information and the second hop attending to the lower-level word unit information based on the decoding and the first attention to the semantic units. Details are presented below.

For the output of the decoder at each time step, it not only attends to the source annotations from the RNN encoder as it usually does but also attends to the semantic unit representations from the MDC. In our model, the decoder output first pays attention to the semantic unit representations from the MDC to figure out the most relevant semantic units and generates a new representation based on the attention. Next, the new representation with both the information from the decoding process as well as the attention to the semantic units attends to the source annotations from the LSTM encoder, so it can extract word-level information from the source text with the guidance of the semantic units, mitigating the problem of irrelevance and redundancy.

To be specific, for the source annotations from the LSTM encoder $h = \{h_1, ..., h_i, ..., h_n\}$ and the semantic unit representations $g = \{g_1, ..., g_i, ..., g_m\}$, the decoder output $s_t$ first attends to the semantic unit representations $g$ and generates a new representation $s'_t$.
% \begin{align}
% s'_t &= Tanh(Ws_t + Uc'_t)\\
% c'_t &= \sum^{m}_{i=1} \alpha_{t,i}h_{i}\\
% \alpha_{t,i} &= \frac{exp(e_{t,i})}{\sum_{j=1}^{m}exp(e_{t,j})}
% \end{align}
Then the new representation $s'_t$ attends to the source annotations $h$ and generates another representation $\tilde{s}_t$ following the identical attention mechanism as mentioned above. In the final step, the model generates $o_t$ for the prediction of $y_t$, where:
\begin{align}
o_t &= s'_t \oplus \tilde{s}_t
\end{align}

For comparison, we also propose another type of attention called ``additive attention'', whose experimental results are in the ablation test. In this mechanism, instead of paying attention to the two types of representations step by step as mentioned above, the output of the LSTM decoder $s_t$ attends to the semantic unit representations $g$ and the source annotations $h$ respectively to generate $s'_t$ and $\tilde{s}_t$, which are finally added element-wisely for the final output $o_t$.

\section{Experiment Setup}
\label{experiment}

In the following, we introduce the datasets and our experiment settings as well as the baseline models that we compare with.

\subsection{Datasets and Preprocessing}
\noindent\textbf{Reuters Corpus Volume I (RCV1-v2)\footnote{\url{http://www.ai.mit.edu/projects/jmlr/papers/volume5/lewis04a/lyrl2004_rcv1v2_README.htm}}:} 
The dataset \citep{rcv1} consists of more than 800k manually categorized newswire stories by Reuters Ltd. for research purpose, where each story is assigned with multiple topics. The total number of topics is 103. To be specific, the training set contains around 802414 samples, while the development set and test set contain 1000 samples respectively. We filter the samples whose lengths are over 500 words in the dataset, which removes about 0.5\% of the samples in the training, development and test sets. The vocabulary size is set to 50k words. Numbers as well as out-of-vocabulary words are masked by special tokens ``\#'' and ``UNK''. For label permutation, we apply the descending order by frequency following \citet{cnn}.

\noindent\textbf{Ren-CECps:} The dataset is a sentence corpus collected from Chinese blogs, annotated with 8 emotion tags \emph{anger}, \emph{anxiety}, \emph{expect}, \emph{hate}, \emph{joy}, \emph{love}, \emph{sorrow} and \emph{surprise} as well as 3 polarity tags \emph{positive}, \emph{negative} and \emph{neutral}. The dataset contains 35096 sentences for multi-label text classification. We apply preprocessing for the data similar to that for the RCV1-v2, which are filtering samples of over 500 words, setting the vocabulary size to 50k and applying the descending order by frequency for label permutation.

\subsection{Experiment Settings}
We implement our experiments in PyTorch on an NVIDIA 1080Ti GPU. In the experiments, the batch size is set to 64, and the embedding size and the number of units of hidden layers are 512. We use Adam optimizer \citep{DBLP:journals/corr/KingmaB14} with the default setting $\beta_{1}=0.9$, $\beta_{2}=0.999$ and $\epsilon=1\times10^{-8}$. The learning rate is initialized to $0.0003$ based on the performance on the development set, and it is halved after every epoch of training. Gradient clipping is applied with the range [-10, 10].

Following the previous studies \citep{ml_knn, cnn-rnn}, we choose hamming loss and micro-${\rm F_1}$ score to evaluate the performance of our model. Hamming loss refers to the fraction of incorrect prediction \citep{hamming_loss}, and micro-${\rm F_1}$ score refers to the weighted average ${\rm F_1}$ score. For reference, the micro-precision as well as micro-recall scores are also reported. To be specific, the computations of Hamming Loss (HL) micro-${\rm F_1}$ score are illustrated below:
\begin{align}
HL &= \frac{1}{L} \sum \mathbb{I}(y \neq \hat{y})\\
microF_1 &= \frac{\sum_{j=1}^{L}2tp_j}{\sum_{j=1}^{L}2tp_j+fp_j+fn_j}
\end{align}
where $tp_j$, $fp_j$ and $fn_j$ refer to the number of true positive examples, false positive examples and false negative examples respectively.

\subsection{Baseline Models}

\begin{table}[tb]
		\centering
%         \footnotesize
    	\setlength{\tabcolsep}{5.7pt}
		\begin{tabular}{ l | c c c c }
		\hline
		\multicolumn{1}{ l |}{\textbf{Models}} &
		\multicolumn{1}{c}{\textbf{HL(-)}} & 
		\multicolumn{1}{c}{\textbf{P(+)}} &  
        \multicolumn{1}{c}{\textbf{R(+)}} &
		\multicolumn{1}{c}{\textbf{F1(+)}}   \\  \hline
        BR & 0.0086 & 0.904 & 0.816 & 0.858\\
		CC & 0.0087  & 0.887 & 0.828 & 0.857\\
        LP & 0.0087  & 0.896 & 0.824 & 0.858 \\
        CNN & 0.0089  & \textbf{0.922} & 0.798 & 0.855 \\
		CNN-RNN & 0.0085 & 0.889 & 0.825 & 0.856 \\ \hline
        S2S & 0.0082 & 0.883 & 0.849 & 0.866\\
        S2S+Attn & 0.0081 & 0.889 & 0.848 & 0.868\\ \hline
        Our Model & \textbf{0.0072}  & 0.891 & \textbf{0.873} & \textbf{0.882}\\ \hline
		\end{tabular}
		\caption{Performance on the RCV1-V2 test set. HL, P, R, and F1 denote hamming loss, micro-precision, micro-recall and micro-${\rm F_1}$, respectively ($p<0.05$).}
		\label{tab_rcv1}
\end{table}

In the following, we introduce the baseline models for comparison for both datasets.

\begin{itemize}
\item \textbf{Binary Relevance (BR)} \citep{br} transforms the MLC task into multiple single-label classification problems.
\item \textbf{Classifier Chains (CC)} \citep{cc} transforms the MLC task into a chain of binary classification problems to model the correlations between labels.
\item \textbf{Label Powerset (LP)} \citep{lp} creates one binary classifier for every label combination attested in the training set.
\item \textbf{CNN} \citep{cnn} uses multiple convolution kernels to extract text feature, which is then input to the linear transformation layer followed by a sigmoid function to output the probability distribution over the label space.
\item \textbf{CNN-RNN} \citep{cnn-rnn} utilizes CNN and RNN to capture both global and local textual semantics and model label correlations.
\item \textbf{S2S} and \textbf{S2S+Attn} \citep{seq2seq,attention} are our implementation of the RNN-based sequence-to-sequence models without and with the attention mechanism respectively.
\end{itemize}

\section{Results and Discussion}

\begin{table}[tb]
		\centering
%         \footnotesize
    	\setlength{\tabcolsep}{5.7pt}
		\begin{tabular}{ l | c c c c }
		\hline
		\multicolumn{1}{ l |}{\textbf{Models}} &
		\multicolumn{1}{c}{\textbf{HL(-)}} & 
		\multicolumn{1}{c}{\textbf{P(+)}} &  
        \multicolumn{1}{c}{\textbf{R(+)}} &
		\multicolumn{1}{c}{\textbf{F1(+)}}   \\  \hline
        BR & \textbf{0.1663} & \textbf{0.649} & 0.472 & 0.546 \\
		CC & 0.1828 & 0.572 & 0.551 & 0.561 \\
        LP & 0.1902 & 0.556 & 0.517 & 0.536 \\
        CNN & 0.1726 & 0.628 & 0.512 & 0.565 \\
		CNN-RNN & 0.1876 & 0.576 & 0.538 & 0.556 \\ \hline
        S2S & 0.1814 & 0.587 & 0.571 & 0.579\\
        S2S+Attn & 0.1793 & 0.589 & 0.573 & 0.581\\\hline
        Our Model &  0.1782 & 0.593 & \textbf{0.585} & \textbf{0.590}\\ \hline
		\end{tabular}
		\caption{Performance of the models on the Ren-CECps test set. HL, P, R, and F1 denote hamming loss, micro-precision, micro-recall and micro-${\rm F_1}$, respectively ($p < 0.05$).}
		\label{tab_ren}
\end{table}

In the following sections, we report the results of our experiments on the RCV1-v2 and Ren-CECps. Moreover, we conduct an ablation test and the comparison with models with hierarchical models with the deterministic setting of sentence or phrase, to illustrate that our model with learnable semantic units possesses a clear advantage over the baseline models. Furthermore, we demonstrate that the higher-level representations are useful for the prediction of labels of low frequency in the dataset so that it can ensure that the model is not strictly learning the prior distribution of the label sequence.

\subsection{Results}
We present the results of our implementations of our model as well as the baselines on the RCV1-v2 on Table~\ref{tab_rcv1}. From the results of the conventional baselines, it can be found that the classical methods for multi-label text classification still own competitiveness compared with the machine-learning-based and even deep-learning-based methods, instead of the Seq2Seq-based models. Regarding the Seq2Seq model, both the S2S and the S2S+Attn achieve improvements on the dataset, compared with the baselines above. However, as mentioned previously, the attention mechanism does not play a significant role in the Seq2Seq model for multi-label text classification. By contrast, our proposed mechanism, which is label-classification-oriented, can take both the information of semantic units and that of word units into consideration. Our proposed model achieves the best performance in the evaluation of Hamming loss and micro-${\rm F_1}$ score, which reduces 9.8\% of Hamming loss and improves 1.3\% of micro-${\rm F_1}$ score, in comparison with the S2S+Attn. 

We also present the results of our experiments on Ren-CECps. Similar to the models' performance on the RCV1-v2, the conventional baselines except for Seq2Seq models achieve lower performance on the evaluation of micro-$\rm {F_1}$ score compared with the Seq2Seq models. Moreover, the S2S and the S2S+Attn still achieve similar performance on micro-$\rm {F_1}$ on this dataset, and our proposed model achieves the best performance with the improvement of 0.009 micro-$\rm {F_1}$ score. An interesting finding is that the Seq2Seq models do not possess an advantage over the conventional baselines on the evaluation of Hamming Loss. We observe that there are fewer labels in the Ren-CECps than in the RCV1-v2 (11 and 103). As our label data are reordered according to the descending order of label frequency, the Seq2Seq model is inclined to learn the frequency distribution, which is similar to a long-tailed distribution. However, regarding the low-frequency labels with only a few samples, their amounts are similar, whose distributions are much more uniform than that of the whole label data. It is more difficult for the Seq2Seq model to classify them correctly while the model is approximating the long-tailed distribution compared with the conventional baselines. As Hamming loss reflects the average incorrect prediction, the errors in classifying into low-frequency labels will lead to a sharper increase in Hamming Loss, in comparison with micro-$\rm {F_1}$ score.

% \begin{table}[tb]
% 		\centering
% %         \footnotesize
%     	\setlength{\tabcolsep}{5.7pt}
% 		\begin{tabular}{ l | c c c c }
% 		\hline
% 		\multicolumn{1}{ l |}{\textbf{Models}} &
% 		\multicolumn{1}{c}{\textbf{HL(-)}} & 
% 		\multicolumn{1}{c}{\textbf{P(+)}} &  
%         \multicolumn{1}{c}{\textbf{R(+)}} &
% 		\multicolumn{1}{c}{\textbf{F1(+)}}   \\  \hline
%         BR & 0.0316 & 0.644 & 0.648 & 0.646 \\
% 		CC & 0.0306 & 0.657 & 0.651 & 0.654 \\
%         LP & 0.0312  & 0.662 & 0.608 & 0.634 \\
%         CNN & 0.0256 & \textbf{0.849} & 0.545 & 0.664 \\
% 		CNN-RNN & 0.0278 & 0.718 & 0.618 & 0.664 \\ \hline
%         S2S & 0.0247 & 0.748 & 0.643 & 0.689\\
%         S2S+Attn & 0.0255 & 0.743 & 0.646 & 0.691\\\hline
%         Our Model &   &  &  & \textbf{0.698}\\ \hline
% 		\end{tabular}
% 		\caption{Performance on the RENcorpus test set. HL, P, R, and F1 denote hamming loss, micro-precision, micro-recall and micro-${\rm F_1}$, respectively. The symbol “+” indicates that the higher the value is, the better the model performs. The symbol “-” is the opposite.}
% 		\label{tab_ren}
% \end{table}

\subsection{Ablation Test}

\begin{table}[tb]
		\centering
%         \footnotesize
    	\setlength{\tabcolsep}{5.7pt}
		\begin{tabular}{ l | c c c c }
		\hline
		\multicolumn{1}{ l |}{\textbf{Models}} &
		\multicolumn{1}{c}{\textbf{HL(-)}} & 
		\multicolumn{1}{c}{\textbf{P(+)}} &  
        \multicolumn{1}{c}{\textbf{R(+)}} &
		\multicolumn{1}{c}{\textbf{F1(+)}}   \\  \hline
        w/o attention & 0.0082 & 0.883 & 0.849 & 0.866\\
        attention & 0.0081 & 0.889 & 0.848 & 0.868\\
        MDC & 0.0074  & 0.889 & 0.871 & 0.880 \\
        additive & 0.0073 & 0.888 & 0.871 & 0.879 \\
		hybrid & \textbf{0.0072} & \textbf{0.891} & \textbf{0.873} & \textbf{0.882} \\ \hline
		\end{tabular}
		\caption{Performance of the models with different attention mechanisms on the RCV1-V2 test set. HL, P, R, and F1 denote hamming loss, micro-precision, micro-recall and micro-${\rm F_1}$, respectively ($p<0.05$). }
		\label{ablation}
\end{table}

To evaluate the effects of our proposed modules, we present an ablation test for our model. We remove certain modules to control variables so that their effects can be fairly compared. To be specific, besides the evaluation of the conventional attention mechanism mentioned in Section~\ref{problem}, we evaluate the effects of hybrid attention and its modules. We demonstrate the performance of five models with different attention implementation for comparison, which are model without attention, one with only attention to the source annotations from LSTM, one with only attention to the semantic unit representations from the MDC, one with the attention to both the source annotations and semantic unit representations (additive) and hybrid attention, respectively. Therefore, the effects of each of our proposed modules, including MDC and hybrid attention, can be evaluated individually without the influence of the other modules.

Results in Table~\ref{ablation} reflect that our model still performs the best in comparison with models with the other types of attention mechanism. Except for the insignificant effect of the conventional attention mechanism mentioned above, it can be found that the high-level representations generated by the MDC contribute much to the performance of the Seq2Seq model for multi-label text classification, which improves about 0.9 micro-$\rm F_1$ score. Moreover, simple additive attention mechanism, which is equivalent to the element-wise addition of the representations of MDC and those of the conventional mechanism, achieves similar performance to the single MDC, which also demonstrates that conventional attention mechanism in this task makes little contribution. As to our proposed hybrid attention, which is a relatively complex combination of the two mechanisms, can improve the performance of MDC. This shows that although conventional attention mechanism for word-level information does not influence the performance of the SeqSeq model significantly, the hybrid attention which extracts word-level information based on the generated high-level semantic information can provide some information about important details that are relevant to the most contributing semantic units.

\subsection{Comparison with the Hierarchical Models}

\begin{table}[tb]
		\centering
%         \footnotesize
    	\setlength{\tabcolsep}{5.7pt}
		\begin{tabular}{ l | c c c c }
		\hline
		\multicolumn{1}{ l |}{\textbf{Models}} &
		\multicolumn{1}{c}{\textbf{HL(-)}} & 
		\multicolumn{1}{c}{\textbf{P(+)}} &  
        \multicolumn{1}{c}{\textbf{R(+)}} &
		\multicolumn{1}{c}{\textbf{F1(+)}}   \\  \hline
        Hier-5 & 0.0075 & 0.887 & 0.869 & 0.878 \\
        Hier-10 & 0.0077 & 0.883 & 0.873 & 0.878 \\
        Hier-15 & 0.0076 & 0.879 & 0.879 & 0.879 \\
        Hier-20 & 0.0076 & 0.876 & \textbf{0.881} & 0.878 \\
		Our model & \textbf{0.0072} & \textbf{0.891} & 0.873 & \textbf{0.882}\\ \hline
		\end{tabular}
		\caption{Performance of the hierarchical model and our model on the RCV1-V2 test set. Hier refers to hierarchical model, and the subsequent number refers to the length of sentence (word) for sentence-level representations ($p<0.05$).}
		\label{heuristic}
\end{table}

Another method that can extract high-level representations is a heuristic method that manually annotates sentences or phrases first and applies a hierarchical model for high-level representations. To be specific, the method does not only apply an RNN encoder to the word representations but also to sentence representations. In our reimplementation, we regard the representation from the LSTM encoder at the time step of the end of each sentence as the sentence representation, and we implement another LSTM on top of the original encoder that receives sentence representations as input so that the whole encoder can be hierarchical. We implement the experiment on the dataset RCV1-v2. As there is no sentence marker in the dataset RCV1-v2, we set a sentence boundary for the source text and we apply a hierarchical model to generate sentence representations.

Compared with our proposed MDC, the hierarchical model for the high-level representations is relatively deterministic since the sentences or phrases are predefined manually. However, our proposed MDC learns the high-level representations through dilated convolution, which is not restricted by the manually-annotated boundaries. Through the evaluation, we expect to see if our model with multi-level dilated convolution as well as hybrid attention can achieve similar or better performance than the hierarchical model. Moreover, we note that the number of parameters of the hierarchical model is more than that of our model, which are 47.24M and 45.13M respectively. Therefore, it is obvious that our model does not possess the advantage of parameter number in the comparison.

We present the results of the evaluation on Table~\ref{heuristic}, where it can be found that our model with fewer parameters still outperforms the hierarchical model with the deterministic setting of sentence or phrase. Moreover, in order to alleviate the influence of the deterministic sentence boundary, we compare the performance of different hierarchical models with different boundaries, which sets the boundaries at the end of every 5, 10, 15 and 20 words respectively. The results in Table~\ref{heuristic} show that the hierarchical models achieve similar performances, which are all higher than the performances of the baselines. This shows that high-level representations can contribute to the performance of the Seq2Seq model on the multi-label text classification task. Furthermore, as these performances are no better than that of our proposed model, it can reflect that the learnable high-level representations can contribute more than deterministic sentence-level representations, as it can be more flexible to represent information of diverse levels, instead of fixed phrase or sentence level.

\begin{figure}[tb]
\centering
\begin{tikzpicture}
\selectcolormodel{RGB}
\begin{axis}[legend pos=north east, xlabel={Ranking of the most frequent label}, ylabel = {micro-$\rm {F_1}$ (\%)}, xmin=0, xmax=60, ymin=60.0, ymax=90.0, xtick=data, ytick={40.0,50.0,60.0,70.0,80.0,90.0}]
\addlegendentry{MDC+Hybrid}
\addplot 
coordinates{(0, 88.2)(10, 82.9)(20, 78.2)(30, 77.1)(40, 72.3)(50, 69.1)(60, 67.7)};
% \legend{Seq2Seq}

\addlegendentry{w/o MDC+Hybrid}
\addplot 
coordinates{(0, 86.9)(10, 79.7)(20, 75.9)(30, 73.7)(40, 69.3)(50, 62.6)(60, 61.9)};
% \legend{ACA}
\end{axis}
\end{tikzpicture}
\caption{\textbf{Micro-$\rm {F_1}$ scores of our model and the baseline on the evaluation of labels of different frequency.} The x-axis refers to the ranking of the most frequent label in the labels for classification, and the y-axis refers to the micro-$\rm {F_1}$ score performance.} 
\label{lambda}

\end{figure}
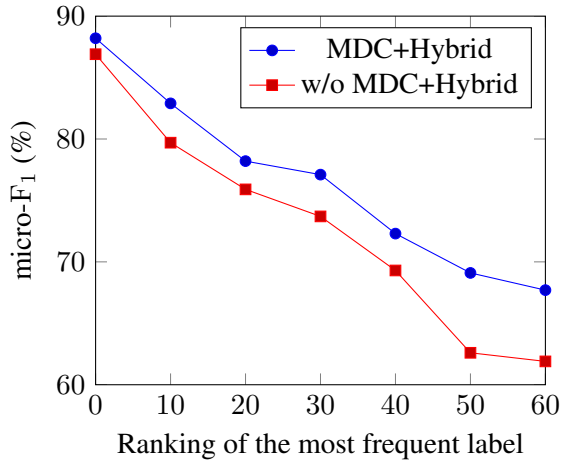

\subsection{Error Analysis}
Another finding in our experiments is that the model's performance on low-frequency label classification is lower than that on high-frequency label classification. This problem is also reflected in our report of the experimental results on the Ren-CECps. The decrease in performance is reasonable since the model is sensitive to the amount of data, especially on small datasets such as Ren-CECps. We also hypothesize that this error comes from the essence of the Seq2Seq model. As the frequency of our label data is similar to a long-tailed distribution and the data are organized by descending order of label frequency, the Seq2Seq model is inclined to model the distribution. As the frequency distribution of the low-frequency labels is relatively uniform, it is much harder for it to model the distribution.

In contrast, as our model is capable of capturing deeper semantic information for the label classification, we believe that it is more robust to the classification of low-frequency labels with the help of the information from multiple levels. We remove the top 10, 20, 30, 40, 50 and 60 most frequent labels subsequently, and we evaluate the performance of our model and the baseline Seq2Seq model on the classification of these labels. Figure~\ref{lambda} shows the results of the models on label data of different frequency. It is obvious that although the performances of both models decrease with the decrease of the label frequency, our model continues to perform better than the baseline on all levels of label frequency. In addition, the gap between the performances of the two models continues to increase with the decrease of label frequency, demonstrating our model's advantage over the baseline on classifying low-frequency labels.

\section{Related Work}
The current models for the multi-label classification task can be classified into three categories: problem transformation methods, algorithm adaptation methods, and neural network models.

Problem transformation methods decompose the multi-label classification task into multiple single-label learning tasks. The BR algorithm \citep{br} builds a separate classifier for each label, causing the label correlations to be ignored. In order to model label correlations, Label Powerset (LP)~\citep{lp} creates one binary classifier for every label combination attested in the training set and Classifier Chains (CC) \citep{cc} connects all classifiers in a chain through feature space. 
% However, the computational efficiency and performance of these methods are challenged by applications with a large number of labels and samples.

Algorithm adaptation methods adopt specific learning algorithms to the multi-label classification task without requiring problem transformations. \citet{ml_dt} constructed decision tree based on multi-label entropy to perform classification. \citet{rank_svm} adopted a Support Vector Machine (SVM) like learning system to handle multi-label problem. \citet{ml_knn} utilized the $k$-nearest neighbor algorithm and maximum a posteriori principle to determine the label set of each sample. \citet{ml_5} made ranking among labels by utilizing pairwise comparison. \citet{lili_mlc} used joint learning predictions as features.
% However, most methods can only be used to capture the first or second order label correlations or are computationally intractable in considering high-order label correlations.

% Among ensemble methods, \citet{rakel} broke the initial set of labels into a number of small random subsets and employed the LP algorithm to train a corresponding classifier. \citet{lsps} proposed to construct a label co-occurrence graph and perform community detection to partition the label set. 

Recent studies of multi-label text classification have turned to the application of neural networks, which have achieved great success in natural language processing. \citet{bp_mll} implemented the fully-connected neural networks with pairwise ranking loss function. \citet{nn} changed the ranking loss function to the cross-entropy loss to better the training. \citet{r4} proposed a novel neural network initialization method to treat some neurons as dedicated neurons to model label correlations. \citet{cnn-rnn} incorporated CNN and RNN so as to capture both global and local semantic information and model high-order label correlations. \citep{nam2017} proposed to generate labels sequentially, and \citet{SGM,semene_pred} both adopted the Seq2Seq, one with a novel decoder and one with a soft loss function respectively.

\section{Conclusion}
In this study, we propose our model based on the multi-level dilated convolution and the hybrid attention mechanism, which can extract both the semantic-unit-level information and word-level information. Experimental results demonstrate that our proposed model can significantly outperform the baseline models. Moreover, the analyses reflect that our model is competitive with the deterministic hierarchical models and it is more robust to classifying the low-frequency labels than the baseline.

% In the future, we will attempt to make the generated representations more interpretable, in order that the contribution of the learnable high-level representations to the improvements of our model can be better explained.

\section*{Acknowledgements}
This work was supported in part by National Natural Science Foundation of China (No. 61673028) and the National Thousand Young Talents Program. Xu Sun is the corresponding author of this paper.

\end{CJK}

\bibliography{emnlp2018}
\bibliographystyle{acl_natbib_nourl}

\end{document}